# A FRACTAL DIMENSION BASED OPTIMAL WAVELET PACKET ANALYSIS TECHNIQUE FOR CLASSIFICATION OF MENINGIOMA BRAIN TUMOURS

*Omar S. Al-Kadi*

Department of Informatics, University of Sussex, Brighton, UK

**ABSTRACT**

With the heterogeneous nature of tissue texture, using a single resolution approach for optimum classification might not suffice. In contrast, a multiresolution wavelet packet analysis can decompose the input signal into a set of frequency subbands giving the opportunity to characterise the texture at the appropriate frequency channel. An adaptive best bases algorithm for optimal bases selection for meningioma histopathological images is proposed, via applying the fractal dimension (FD) as the bases selection criterion in a tree-structured manner. Thereby, the most significant subband that better identifies texture discontinuities will only be chosen for further decomposition, and its fractal signature would represent the extracted feature vector for classification. The best basis selection using the FD outperformed the energy based selection approaches, achieving an overall classification accuracy of 91.25% as compared to 83.44% and 73.75% for the co-occurrence matrix and energy texture signatures; respectively.

## 1. INTRODUCTION

Meningiomas are one of the most common type of primary brain tumours in adults, and it occurs twice as frequent in women than men [1]. Although most meningiomas are diagnosed as benign, the occurrence of these tumours in a very sensitive organ renders them very serious and maybe life threatening. Unfortunately, variability in physician's diagnostic decision exists [2]. Meaning there is a risk for incorrectly determining the tissue healthiness state, or in misclassifying the tumour type or grade; which could contribute towards a misleading patients' prognosis. Thus, there is a need for a more reliable technique which can assist physicians for a more accurate meningioma diagnosis.

Medical texture is known to be non-stationary; therefore a multiresolution perspective using wavelets would assist in achieving a better characterisation of tumour images. Wavelet packets (WPs) are a generalised framework of wavelets transform and comprise all possible combination of subbands decomposition. However, it is unwieldy to use all frequency subbands for texture characterisation as not all of them have the same discriminating power, and inclusion of weak subbands would have a negative impact on the classifier's performance. Using an exhaustive search would also be computational expensive as the number of decomposition levels grows higher. Therefore an adaptive approach is required for selection of the basis with prominent discriminating power. The selection criteria can be done either by selecting the best bases from a library of WPs or in a tree-structured approach. Coifman and Wickerhauser proposed the use of entropy as a cost function to choose the best WP basis which gave the most compact representation [3]. Laine and Fan used a two layer network classifier for classifying energy and entropy measures computed from each WP [4]. Saito et al estimated the probability density of each class in each coordinate in the WP and local trigonometric bases, then applied the relative entropy as a distance measure among the densities for selection of the most discriminating coordinates [5]. Rajpoot compared the discrimination energy between the subbands by using four different distance metrics [6]. Another work was based on best clustering bases, wherein clustering basis functions are selected according to their ability to separate the fMRI time series into activated and non-activated clusters [7]. On the other hand, a tree-structured technique for best basis selection was proposed by Chang and Kuo, where only the subbands with the highest energy are selected for further decomposition [8]. Acharyya and Kundu used an *M*-band WP decomposition based on a tree-structured approach [9]. Regarding application of WPs to meningiomas, some used unsupervised learning techniques for training artificial neural networks to classify features derived by WP transform [10]. Also in another two studies, the performance of extracted features using adaptive WP transform was compared to local binary patterns [11] and to co-occurrence methods [12] via a support vector machine classifier.

In this work a different approach for best basis selection for the processes of histopathological meningioma classification is proposed. The fractal dimension (FD) is used for guiding the subband tree-structure decomposition instead of energy which is highly dependent on the subband intensity. The motivation to use such texture measure, besides its scale invariance or the capability to investigate self-similarity, is its surface roughness estimation that can be used to detect variations between meningioma cell nuclei subtypes. Fractal analysis for the purpose of tumour discrimination was proven to be successful in numerous studies related to various medical imaging modalities as in CT [13], X-ray [14], MR [15], and US [16]. This work takes advantage of FD in diagnosing medical texture, and applies it to images acquired by electronic microscopy modality. Also with the large size of the meningioma images (512 x 512 pixels), the tree-structured approach was favoured to reduce computational time in order to explore the full texture characteristics at deeper levels, as an overcomplete dyadic wavelet transform was applied.

## 2. MULTIRESOLUTION VIA WAVELETS

Multiresolution processing gives the advantage of analysing both small and large object characteristics in a single image at several resolutions. The decomposition of the image into multiple resolutions based on small basis functions of varying frequency and limited duration called wavelets was first introduced by Mallat [17]. The wavelet analysis approach can be regarded as the scale *j* and translation $k$ of a basic function to cover the spatial-frequency domain. A one-dimensional decomposition of a function $f(x) \in L^2(\mathbb{R})$ relative to scaling $\varphi(x)$ and wavelet function $\psi(x)$, where $\varphi_{j,k}(x) = 2^{j/2}\varphi(2^j x - k)$ and $\psi_{j,k}(x) = 2^{j/2}\psi(2^j x - k)$ for all $j, k \in \mathbb{Z}$ and $\varphi(x)$ and $\psi(x) \in L^2(\mathbb{R})$, can be written in the following expansion:

$$f(x) = \sum_k c_{j_0}(k)\,\varphi_{j_0,k}(x) + \sum_{j=j_0}^{\infty}\sum_k d_j(k)\psi_{j,k}(x) \quad (1)$$

where $j_0$ is an arbitrary starting scale, and the expansion coefficients $c_{j_0}(k)$ and $d_j(k)$ are determined by

$$c_{j_0}(k) = \langle f(x), \varphi_{j_0,k}(x)\rangle = \int f(x)\varphi_{j_0,k}(x)\,dx \quad (2)$$
$$d_j(k) = \langle f(x), \psi_{j,k}(x)\rangle = \int f(x)\,\psi_{j,k}(x)\,dx \quad (3)$$

where $\{\varphi(x), \psi(x)\}$ are mutually orthogonal functions and $\langle,\rangle$ is the inner product operator. $\varphi(t)$ satisfies the dilation equation $\varphi(x) = \sqrt{2}\sum_k h_0(k)\,\varphi(2x - k)$ with $h_0(k)$ denoting scaling filter, while $\psi(x)$ satisfies the wavelet equation $\psi(x) = \sqrt{2}\sum_k h_1(k)\,\psi(2x - k)$ with $h_1(k)$ denoting wavelet filter. These two filters need to satisfy certain conditions for the set of basis wavelet functions to be unique and orthonormal [17, 18]. By decomposing the signal's approximation coefficients $d_j(k)$ as well, the wavelet transform can be extended in the middle and high frequency channels (LH, HL and HH bands) and not only in the low frequency channels (LL-band). This provides a better partitioning of the spatial-frequency domain, which is known as WP transform [3]. Features in some textures would be more prevalent in the higher frequency channels, thus WPs would give the high frequency structures in an image an equal opportunity for investigation of possible interesting information. Herein, we are more concerned with a better representation of the texture characteristics at each decomposition and not compression, therefore an overcomplete tree-structured wavelet via an 8-tap Daubechies filter was used by holding the size of the transformed image the same as the original image.

**Table I** Fractal dimensions for each corresponding meningioma fibroblastic subtype wavelet subband.

| Resolution | $W_{LL}$ | $W_{LH}$ | $W_{HL}$ | $W_{HH}$ |
|---|---|---|---|---|
| level 1 | 2.5038 | 2.6797 | 2.7105 | 2.7897 |
| level 2 | 2.9346 | 2.8918 | 2.8994 | 2.8239 |
| level 3 | 2.9585 | 2.9655 | 2.9669 | 2.9722 |
| level 4 | 2.9877 | 2.9857 | 2.9860 | 2.9838 |
| level 5 | 2.9930 | 2.9937 | 2.9939 | 2.9945 |
| level 6 | 2.9975 | 2.9972 | 2.9973 | 2.9970 |
| level 7 | 2.9986 | 2.9987 | 2.9987 | 2.9988 |
| level 8 | 2.9994 | 2.9994 | 2.9994 | 2.9994 |

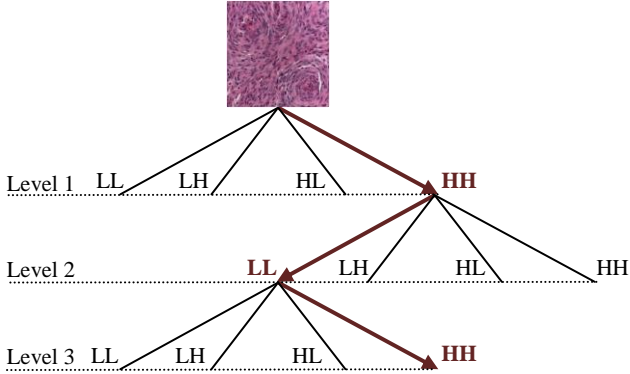

**Fig. 1.** Wavelet quad-tree structure for first 3 levels in Table I.

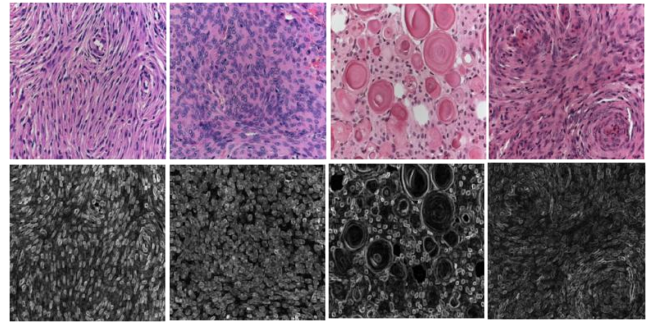

**Fig. 2.** Upper row (left-right) is the meningioma fibroblastic, meningothelial, psammomatous and transitional subtypes. Second row is the corresponding grey-level segmented cell nuclei general structure.

### 3. SUBBAND SELECTION OPTIMISATION

The FD signatures are estimated for all subbands at each level of WP decomposition. Then the subband with the highest FD is selected for further decomposition. There are several fractal models used to estimate the FD; the fractal Brownian motion which is the mean absolute difference of pixel pairs as a function of scale as shown in (4) was adopted [19].

$$E(\Delta I) = K \Delta r^H \quad (4)$$

where $\Delta I = |I(x_2, y_2)| - |I(x_1, y_1)|$ is the mean absolute difference of pixel pairs; $\Delta r = \sqrt{[(x_2 - x_1) + (y_2 - y_1)]}$ is the pixel pair distances; $H$ is called the Hurst coefficient; and $K$ is a constant. The FD can be then estimated by plotting both sides of (4) on a log-log scale and $H$ will represent the slope of the curve that is used to estimate the FD as: FD = 3 – H. For example, Table I lists the estimated FD values for each subband at each decomposition level, where the $W_{HH}$ subband which had the highest FD value for first resolution level was the chosen basis for the second decomposition level, and so on. The quad-tree structure for the first three decomposition levels is shown in Fig. 1.

At the end of the feature extraction stage, a feature vector $W_{FD} = (f_1^i, f_2^i, \ldots f_j^i)$ consisting of all selected subbands FD signatures $f$ to a certain decomposition level $j$ will be produced for each of the meningioma subimages $i$. In order to save processing time and when the difference in-between the FD signatures become less significant, a designated threshold $\lambda$ would reduce the dimensionality of the extracted feature vector. By that, unnecessary decompositions are avoided which could have a negative effect on the classifier's performance. This can be expressed as if the condition $(\forall f_j^i \in W_{FD}) \leq \lambda$ is satisfied, then the decomposition should terminate. Therefore, the FD signatures' absolute difference $D_f = |f_1^j - f_2^j|$ between all four wavelets subbands ($W_{LL}, W_{LH}, W_{HL}$ and $W_{HH}$) for a certain resolution level needs to be less than or equal to $\lambda$ (empirically choosing $\lambda = 0.05$ for psammomatous and 0.012 for the other subtypes) before decomposition terminates.

**Table II** Wavelet packet decomposition using maximum fractal dimension signature for best basis selection.

| Resolution | Meningioma subtype | | | | Total Accuracy |
|---|---|---|---|---|---|
| | Fib. | Men. | Psa. | Tra. | |
| level 1 | 65.00 | 91.25 | 73.75 | 43.75 | 68.44% |
| level 2 | 82.50 | 91.25 | **95.00** | 86.25 | 88.75% |
| level 3 | 83.75 | **92.50** | 93.75 | **91.25** | 90.31% |
| level 4 | **86.25** | 86.25 | 93.75 | 88.75 | 88.75% |
| level 5 | 35.00 | 91.25 | 85.00 | 70.00 | 70.31% |
| level 6 | 75.00 | 82.50 | 91.25 | 27.50 | 69.06% |
| level 7 | 72.50 | 82.50 | 90.00 | 28.75 | 68.44% |
| level 8 | 47.50 | 75.00 | 87.5 | 40.00 | 62.50% |

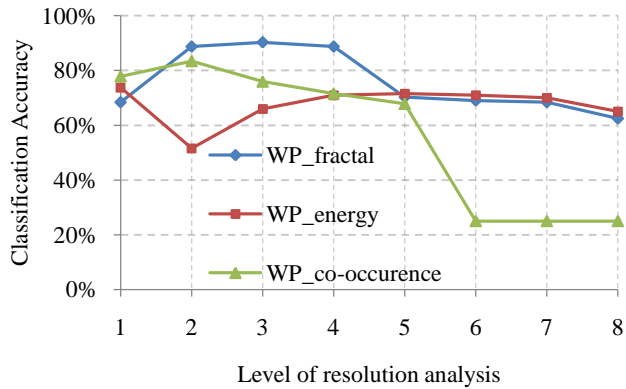

**Fig. 3.** Multiresolution level wavelet packet comparison of meningioma classification accuracy based on fractal dimension, energy and co-occurrence texture signatures.

## 4. TEXTURE SIGNATURE CLASSIFICATION

A naïve Bayesian classifier (NBC) was applied for meningioma classification. A NBC in supervised learning can achieve optimal accuracy if all attributes are independent given the class. Despite the fact that this condition might not be frequent in practice, this fast and simple classifier was reported to perform well even with the presence of strong attribute dependence [20]. According to Bayesian theory and after assuming conditional independence of attributes values, NBC can be represented as

$$P(C_i/X) = \arg max \left( \frac{P(C_i) \prod_{j=1}^{n} P(X_j/C_i)}{\sum_{i=1}^{k} P(C_i) \prod_{j=1}^{n} P(X_j/C_i)} \right) \quad (5)$$

where $P(C_i/X)$ is the *a posteriori* probability of assigning class $i$ given feature vector $X$, $P(X/C_i)$ is the probability density function of $X$ within the $i$th class $C_i$ for a total number of $k$ classes, $P(C_i)$ is the *a priori* probability of class $C_i$. A leave-one-out approach [21] was applied to validate the classification results, which is done by designing the classifier using ($n$-1) samples and then evaluated on the remaining set-aside sample. This process is repeated $n$ times covering all possible unique sets of other samples. Thereby an unbiased estimation is achieved although the performance is sometimes overestimated.

## 5. EXPERIMENTAL RESULTS

### 5.1. Image pre-processing

A data-set of 320 grade I meningioma EM images with a 512 × 512 pixels resolution was used in this study. Each of the four different meningioma subtypes (fibroblastic, meningothelial, psammomatous and transitional) are equally represented by 80 images in the data-set. The subtypes were first segmented prior to feature extraction, see Fig. 2. This is done by investigating the separability of the RGB colour channels before selecting the appropriate colour channel for cell nuclei segmentation, followed by applying the morphological gradient to extract the general structure and elimination of any possible tissue cracks that may occurred during biopsy preparation procedure. This assists in highlighting the size and orientation of the cell nuclei structure, which would reflect on the quality of the texture signatures to be extracted from each subband.

### 5.2. Classification Results

The classification performance with up to eight levels of resolution using the FD signature for best basis selection (BBS$_{FD}$) is shown in Table II, where a threshold value for the FD signature was not used to stop the decomposition. The best classification accuracy of 90.31% was achieved at the third level of decomposition. Alternatively, using the appropriate threshold as discussed in the subband optimisation section, the decomposition should terminate when there is no significant difference between the FD signatures ─ highlighted in bold ─ giving a slightly improved overall accuracy of 91.25%.

A comparison is also performed to evaluate the performance of the BBS$_{FD}$ approach with two other statistical methods. The BBS$_{FD}$ model based method suggested in this paper used the FD signatures to guide the WP tree-structured expansion in order to construct a feature vector of the subbands having the highest FD signatures. On the other hand, the statistical approaches used the highest energy for best basis selection process, where the first method (abbreviated BBS$_E$) simply employed the computed highest energies of the subbands as signatures, and the second method (abbreviated BBS$_{CM}$) extracted the co-occurrence matrix correlation, entropy and energy (with unit distance $\delta = 1$ and four orientation $\theta = 0°$, 45°, 90° and 135°) as second order statistical signatures for classification. The three subband decomposition approaches were also run at up to eight levels of resolution, and the corresponding classification accuracy is determined at each level. It is evident from Fig. 3 that the BBS$_{FD}$ fractal approach outperformed the others, where the BBS$_{CM}$ and BBS$_E$ approaches achieved a maximum overall classification accuracy of 83.44% and 73.75%; respectively.

## 6. INTERPRETATION AND CONCLUSION

Results showed that maximum classification accuracy was reached within two to four resolution decompositions ─ depending on the subtype ─ before starting to degrade. This was expected as the FD measures would give a reliable estimation to a certain level of resolution, whereas the more levels are decomposed the less details remains for the FD to measure. Thus, determining the appropriate resolution level is not only important to save computational time but also to improve the quality of the extracted subband features. The subband discriminating power was considered trivial by measuring the FD signature absolute difference between all subbands after empirically specifying a threshold, which specifies how deep the image resolution can be probed. This is equivalent to excluding FD signatures equal to or above 2.985, considering them as nearer to noise rather than a meaningful roughness estimation of the surface. Moreover, comparing the suggested $BBS_{FD}$ technique with the statistical $BBS_{CM}$ and $BBS_E$ techniques, a significant improvement in the classification accuracy was achieved by 7.81% and 17.50% if a threshold was used for determination of decomposition levels and a 6.87% and 16.56% improvement if a fixed level of decomposition was applied (three levels for this case); respectively.

The reason for meningioma subtypes not having their optimum classification performance at an equal level can be referred to the cell nuclei denseness variation between subtypes. Depending on the subtype, denseness here means the size and number of cell nuclei existing in a biopsy and whether they overlap or not. Subtypes having many small size cell nuclei would expect to represent a rougher surface as compared to small overlapping or large size ones (i.e. fewer edges to detect). Therefore after the segmentation process the general structure of the cell nuclei distribution in each subtype is what remains, and the segmented images with more edges would be regarded richer with texture information. For example, higher resolution levels would be more appropriate to analyse psammomatous subtypes which have less texture details (i.e. cell structure is less dense as compared to other subtypes, which required tweaking λ from 0.012 to 0.05 for decomposition to terminate earlier), while lower levels would be more appropriate for the remaining three other subtypes.

A different approach via measuring the fractal dimension for tree structured wavelet decomposition demonstrated its performance in distinguishing grade I histopathological meningioma images with an improved accuracy as compared to conventional energy based decomposition. The $BBS_{FD}$ relies on revealing texture structure complexity which would better characterising the information situated in the middle and high frequency bands. Also, the appropriate decomposition level would be detected when no more significant difference in-between the subbands exist, saving unnecessary computational operations. Possible future developments would be investigating if the quality of the extracted feature would improve if an *M*-band wavelet transform for subband decomposition to be used and compare the performance with redundant decomposition techniques. Furthermore, testing this technique on other grades of meningioma or different types of brain tumours would assist in benchmarking the technique's performance.

## ACKNOWLEDGMENT

The author would like to thank Dr. Volkmar Hans from the Institute of Neuropathology, Bielefeld, Germany for provision of the meningioma data-set, and the anonymous reviewers for their constructive comments.

## 7. REFERENCES

[1] F. DeMonte, M. R. Gilbert, A. Mahajan, and I. E. McCutcheon *Tumors of the Brain and Spine* 1ed.: Springer, 2007.
[2] A. D. Ramsay, "Errors in histopathology reporting: detection and avoidance," *Histopathology,* vol. 34, pp. 481-490, Jun 1999.
[3] R. R. Coifman and M. V. Wickerhauser, "Entropy-based algorithms for best basis selection," *IEEE Trans. Inform. Theory,* vol. 38, pp. 713-718, Mar 1992.
[4] A. Laine and J. Fan, "Texture classification by wavelet packet signatures," *IEEE Trans. Pattern Anal. and Machine Intell.,* vol. 15, pp. 1186-1191, Nov 1993.
[5] N. Saito, R. R. Coifman, F. B. Geshwind, and F. Warner, "Discriminant feature extraction using empirical probability density estimation and a local basis library," *Pattern Recognition,* vol. 35, pp. 2841-2852, Dec 2002.
[6] N. Rajpoot, "Local discriminant wavelet packet basis for texture classification," in Proc. *SPIE Wavelets X, San Diego,* California, 2003, pp. 774-783.
[7] F. G. Meyer and J. Chinrungrueng, "Analysis of event-related fMRI data using best clustering bases," *IEEE Trans. Med. Imaging,* vol. 22, pp. 933-939, Aug 2003.
[8] T. Chang and C. C. J. Kuo, "Texture analysis and classification with tree-structured wavelet transform," *IEEE Trans. Image Processing,* vol. 2, pp. 429-441, 1993.
[9] M. Acharyya, R. K. De, and M. K. Kundu, "Extraction of features using M-band wavelet packet frame and their neuro-fuzzy evaluation for multitexture segmentation," *IEEE Trans. Pattern Anal. and Machine Intell.,* vol. 25, pp. 1639-1644, Dec 2003.
[10] B. Lessmann, T. W. Nattkemper, V. H. Hans, and A. Degenhard, "A method for linking computed image features to histological semantics in neuropathology," *J. Biomed. Inform.,* vol. 40, pp. 631-641, 2007.
[11] H. Qureshi, O. Sertel, N. Rajpoot, R. Wilson, and M. Gurcan, "Adaptive Discriminant Wavelet Packet Transform and Local Binary Patterns for Meningioma Subtype Classification," in *Proc. MICCAI, 2008,* vol. 5242, pp. 196-204, 2008.
[12] H. Qureshi, N. Rajpoot, R. Wilson, T. Nattkemper, and V. Hans, "Comparative analysis of discriminant wavelet packet features and raw image features for classification of meningioma subtypes," in *Proc. MIUA*, UK, 2007.
[13] O. S. Al-Kadi and D. Watson, "Texture Analysis of Aggressive and non-Aggressive Lung Tumor CE CT Images," *IEEE Trans. Biomed. Eng.,* vol. 55, pp. 1822-1830, 2008.
[14] M. E. Mavroforakis, H. V. Georgiou, N. Dimitropoulos, D. Cavouras, and S. Theodoridis, "Mammographic masses characterization based on localized texture and dataset fractal analysis using linear, neural and support vector machine classifiers," *Artificial Intell. in Med.,* vol. 37, pp. 145-162, 2006.
[15] K. M. Iftekharuddin, W. Jia, and R. Marsh, "Fractal analysis of tumor in brain MR images," *Machine Vision and Applications,* vol. 13, pp. 352-362, Mar 2003.
[16] W.-L. Lee, Y. Chen, and K. Hsieh, "Ultrasonic liver tissue classification by fractal feature vector based on M-band wavelet transform," *IEEE Trans. Med. Imaging,* vol. 22, pp. 382-392, 2003.
[17] S. G. Mallat, "A theory for multiresolution signal decomposition - the wavelet representation," *IEEE Trans. Pattern Anal. and Machine Intell.,* vol. 11, pp. 674-693, Jul 1989.
[18] I. Daubechies, "The wavelet transform, time-frequency localization and signal analysis," *IEEE Trans. Inform. Theory,* vol. 36, pp. 961-1005, 1990.
[19] C. C. Chen, J. S. Daponte, and M. D. Fox, "Fractal Feature Analysis and Classification in Medical Imaging," *IEEE Trans. Med. Imaging,* vol. 8, pp. 133-142, 1989.
[20] P. Domingos and M. Pazzani, "On the optimality of the simple Bayesian classifier under zero-one loss," *Machine Learning,* vol. 29, pp. 103-130, Nov-Dec 1997.
[21] A. K. Jain, R. P. W. Duin, and J. C. Mao, "Statistical pattern recognition: A review," *IEEE Trans. Pattern Anal. and Machine Intell.,* vol. 22, pp. 4-37, 2000.